\documentclass[letterpaper, 10 pt, conference]{IEEEtran}  

\IEEEoverridecommandlockouts                              

\usepackage{amsmath} 
\usepackage{amssymb}  
\usepackage{todonotes}
\usepackage{enumitem}
\usepackage{mathtools}
\usepackage{rotating}
\usepackage{multirow}
\usepackage{siunitx}
\usepackage{caption}
\usepackage{subcaption}
\usepackage{float}
\usepackage{hyperref}
\usepackage{pifont}
\usepackage{cite}

\usepackage{booktabs} 

\usepackage{algorithmic} 
\usepackage{algorithm2e}
\graphicspath{{graphics/}}

\usepackage{tabto} 

\usetikzlibrary{matrix, positioning, calc}
\tikzset{
  quadratic/.style={
    to path={
      (\tikztostart) .. controls
      ($#1!1/3!(\tikztostart)$) and ($#1!1/3!(\tikztotarget)$)
      .. (\tikztotarget)
    }
  }
}

\hypersetup{
    colorlinks=true,
    linkcolor=black,
    filecolor=magenta,      
    urlcolor=cyan,
    pdftitle={CoLD Fusion},
    pdfpagemode=FullScreen,
    }

\title{
CoLD Fusion: A Real-time Capable Spline-based Fusion Algorithm for Collective Lane Detection 
}

\author{\IEEEauthorblockN{Jörg Gamerdinger, Sven Teufel, Georg Volk and Oliver Bringmann}
\IEEEauthorblockA{\textit{Department of Computer Science} \\
\textit{University of Tübingen}\\
Tübingen, Germany \\
\{joerg.gamerdinger, sven.teufel, georg.volk, oliver.bringmann\}@uni-tuebingen.de}
}

\begin{document}
\maketitle
\thispagestyle{empty}
\pagestyle{empty}

\begin{abstract}
Comprehensive environment perception is essential for autonomous vehicles to operate safely.
It is crucial to detect both dynamic road users and static objects like traffic signs or lanes as these are required for safe motion planning. However, in many circumstances a complete perception of other objects or lanes is not achievable due to limited sensor ranges, occlusions, and curves. In scenarios where an accurate localization is not possible or for roads where no HD maps are available, an autonomous vehicle must rely solely on its perceived road information.
Thus, extending local sensing capabilities through collective perception using vehicle-to-vehicle communication is a promising strategy that has not yet been explored for lane detection. Therefore, we propose a real-time capable approach for collective perception of lanes using a spline-based estimation of undetected road sections. We evaluate our proposed fusion algorithm in various situations and road types. We were able to achieve real-time capability and extend the perception range by up to 200\%.
\end{abstract}


\section{INTRODUCTION}
\label{sec:intro}
In the European Union, \SI{90}{\percent} of traffic accidents with fatalities can be attributed to human error~\cite{europeanUnion2019}. As a result, research and industry have become increasingly interested in autonomous driving because it has the potential to improve traffic flow and increase road safety.
Autonomous vehicles must overcome various challenges in order to make this contribution, such as having a proper and full perception of their surroundings and having accurate trajectory planning. 

The vehicle-local perception is limited by sensing ranges or occlusion through e.g. buildings and moreover affected by environmental influences like rain~\cite{volk2019towards}. A promising solution to deal with these challenges is collective perception (CP). In CP multiple vehicles perceive their surroundings and transmit information about their state and detected objects via vehicle-to-vehicle (V2V) or vehicle-to-everything (V2X) communication to extend the local perception of an ego vehicle to a more complete environment. CP increases sensor ranges and enhances perception quality as shown by Volk et al.~\cite{volk_environment-aware_2019, volk2021}. Besides the perception of other road users, the perception of lanes is crucial to perform the motion planning without HD maps, which are often not available. The lane detection is affected by environmental influences and occlusion in the same way as the object detection. Increased safety, improved traffic flow, and decreased fuel consumption can all be attributed to extended lane understanding and a resulting extended trajectory planning. However, to the best of our knowledge, the concept of CP has not been applied to lane detection. 

The goal of this work is the investigation and development of a collective lane detection (CoLD) system for convoy driving (short distance) and regular driving with medium distances between vehicles. In Sec.~\ref{sec:related_work}, we present an extract of works with reference to our collective lane detection. A detailed overview of our spline-based approach, including our proposed apex estimation for an improved interpolation is presented in Sec.~\ref{sec:method}. Then, we show the results of our proposed method based on three different scenarios in terms of real-time performance and perceptual accuracy in Sec.~\ref{sec:results}. Finally, in Sec.~\ref{sec:conclusion} we conclude our work and show an outlook to further research.

\section{RELATED WORK}
\label{sec:related_work}

Different representations and frameworks, to work with lanes and maps already exist.
Basically, lanes can be described through mathematical primitives such as lines, arcs, clothoids, and polynomial functions~\cite{dupuis2010opendrive}. Moreover, piecewise polynomial functions, so called splines, are suitable to describe lanes as shown in the works of Wang et al.~\cite{wang2000lane} and Zhao et al.~\cite{zhao2012novel}. Towards polynomial functions a spline avoids the phenomenon of Runge~\cite{runge1901empirische} which generates oscillations while interpolating points.

A comprehensive map format using mathematical primitives is the OpenDrive format by Dupuis et al.~\cite{dupuis2010opendrive}. OpenDrive provides information about the centerline of road sections based on the above mentioned primitives. Lanes are provided in relation to the centerline. The format allows the integration of various additional information such as road surface, speed limits or infrastructural elements like traffic signals. 
A well known but complex format is the Open Street Map (OSM) description by Haklay and Weber~\cite{haklay2008openstreetmap}. The open project collects geo data with millions of registered users and provides maps free to use. A more simple framework to describe lanes and road networks is the Lanelet framework by Bender et al.~\cite{lanelet} with the improvements in Lanelet2 by Poggenhans et al.~\cite{lanelet2}. The framework provides a XML-based description of maps by single lanes which consists of two point lists for left and right lane boundary. Furthermore, Lanelet2 comes along with different modules to incorporate traffic rules, constructing trajectories based on the given road network topology or modules to match coordinates to lanelets. 
The CommonRoad project~\cite{CommonRoad} is a benchmark framework for motion planning which provides different tools to verify motion planning and to convert maps between different map formats like the OpenDrive2Lanelet converter~\cite{opendrive2lanelet}.

Vehicle-local lane detection is investigated by many researchers. Early works such as Wang et al.~\cite{wang2000lane} or Aly~\cite{aly2008real} used image-based edge detection algorithms combined with the hough transformation to detect lane markings. Based on actual benchmarks like CULane~\cite{pan2018SCNN}, it can be observed that the best results are mostly achieved by neural network-based lane detectors. Zou et al.~\cite{zou2019robust} used a combination of convolutional and recurrent neural networks to achieve a precision of about \SI{98}{\percent}. Considering the review of deep neural network lane detectors by Mamun et al.~\cite{mamun2022comprehensive} various algorithms are capable of achieving precision and accuracy scores of \SIrange{95}{97}{\percent}.

\begin{figure}[h!]
    \centering
    \includegraphics[width=0.94\linewidth, page=3, trim=6.5cm 0.5cm 6.5cm 0.7cm, clip]{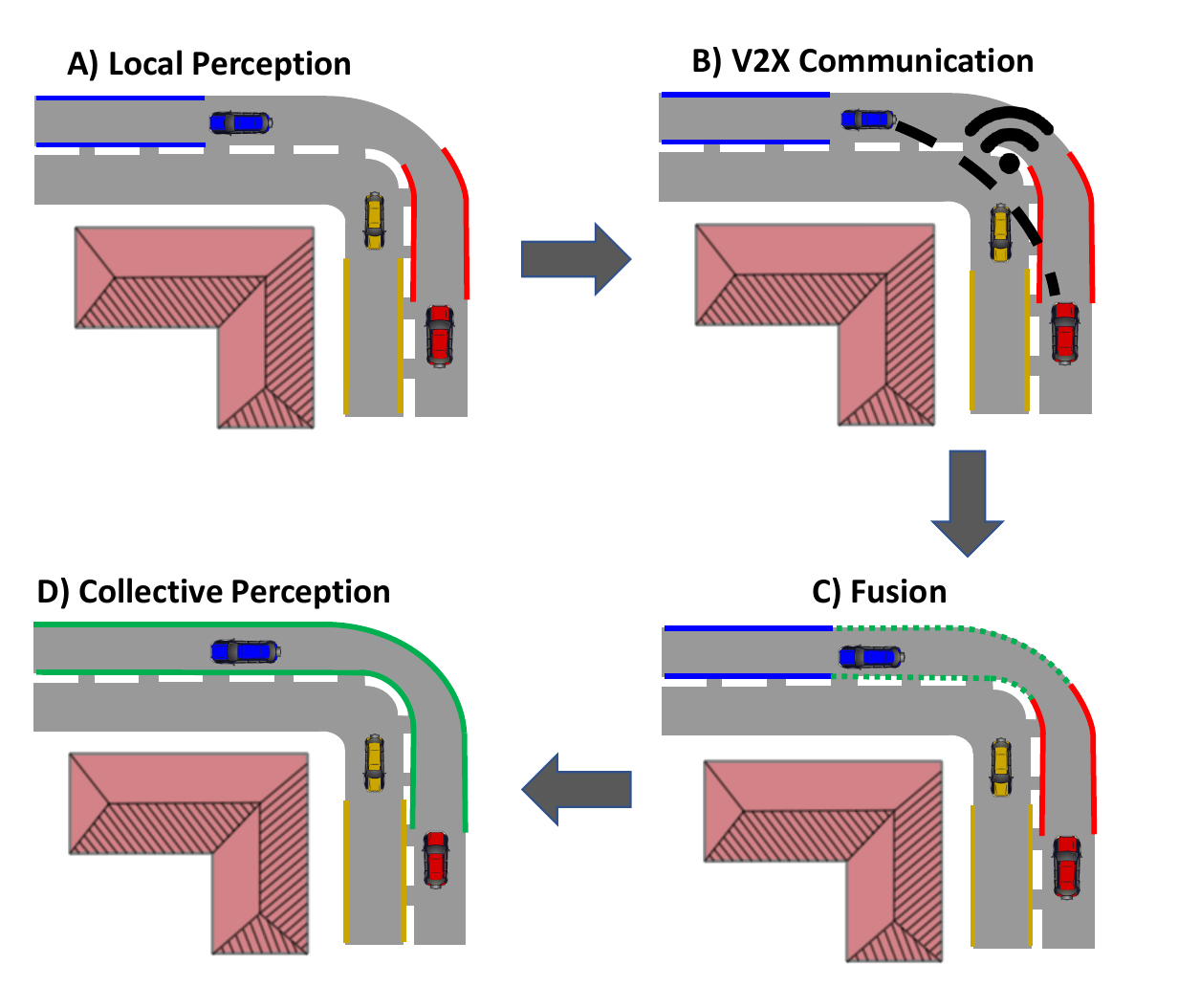}
    \caption{Overview of the collective perception pipeline for CoLD. A) describes a given scenario with local perception, in B) the perceived lanes are transmitted via V2V/V2X communication. C) shows the fusion after a validation and D) the final collectively perceived lane (green). Image adapted from~\cite{volk_environment-aware_2019}.} 
    \label{fig:coop-overview}
\end{figure}

Due to limited sensing capability or occlusions, local perception may not always satisfy sensing range requirements. A promising approach to solve this is the collective perception. A short overview of the CP process adapted to our CoLD system is shown in Fig.~\ref{fig:coop-overview}. 

An early work for CP of road users was presented by Rauch et al.~\cite{rauch2012}. They proposed methods for data fusion and data maintenance in a Car2X-based object perception system. The perception advantage of CP towards local perception under adverse weather was investigated by Volk et al.~\cite{volk_environment-aware_2019}. They show that CP still achieves an average precision of about \SI{24}{\percent}, while the local perception was not capable to perceive any object due to dense fog. Furthermore, Schiegg et al.~\cite{schiegg2020} investigated the performance of the collective perception service with cellular V2X. For the information exchange the European Telecommunications Standards Institute (ETSI) proposed message formats as well as generation rules. The two defined message formats are the \textit{Cooperative Awareness Message} (CAM)~\cite{ETSI-CAM}, containing information about the ego vehicle and the \textit{Collective Perception Message} (CPM) containing information about the ego state as well as the states of the perceived objects. These messages can be seen as work-in-progress standard. The message generation rules have been revised by Delooz et al.~\cite{delooz2020} and Thandavarayan et al.~\cite{thandavarayan2019}. 

\section{COLLECTIVE LANE DETECTION}
\label{sec:method}
\begin{figure}[t]
    \centering
    \includegraphics[width=\linewidth, page=2, trim=4.5cm 8cm 4.5cm 0.5cm, clip]{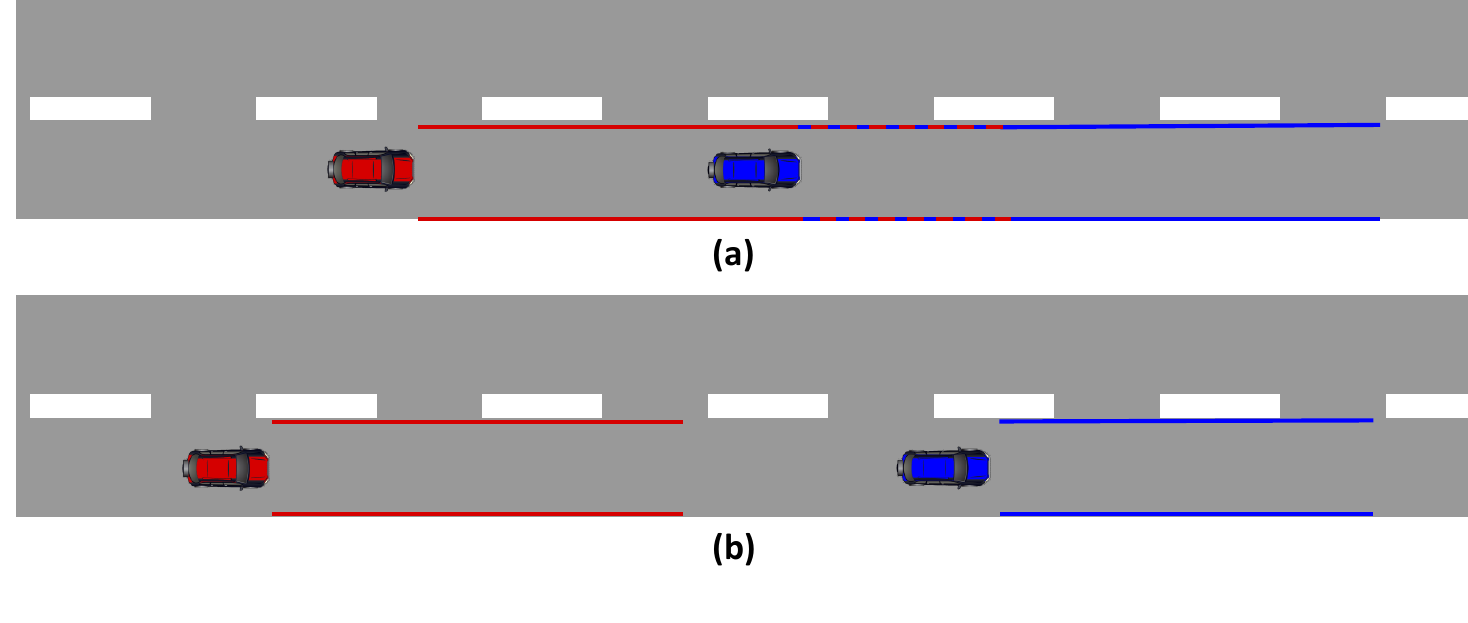}
    \caption{Possible fusion types to distinguish in our CoLD. (a) shows the convoy scenario (referred to as convoy fusion) with a cooperative vehicle (blue) inside of the locally detected lane by the ego vehicle (red). (b) shows the regular driving scenario (referred to as spline fusion) with a detection gap between the local detections of ego and cooperative vehicle.} 
    \label{fig:cld-cases}
\end{figure}

As presented in Sec.~\ref{sec:related_work}, different approaches for vehicle-local lane detection and collective perception of objects exist; we aim to combine both. Therefore, in this section we will propose an approach for the collective lane detection in (a) a convoy which is comparable to platooning but not explicitly planned as well as (b) for regular driving vehicles. The distinction between these two cases is shown in Fig.~\ref{fig:cld-cases}. In this section $l_{\mathrm{ego}}$ refers to the locally perceived lane of the ego vehicle and $l_{\mathrm{coop}}$ refers to the locally perceived lane of the cooperative vehicle. The resulting collectively perceived lane is denoted as $l_{\mathrm{coll}}$. It must be stated that in contrast to the fusion of dynamic objects (e.g. vehicles), possible communication delays can be neglected since they do not affect static objects. Hence, no alignment in time is necessary for our fusion approach. Due to the early stage of development, some assumptions have to be made: we assume an accurate localization and resulting from this an error free transformation between coordinate systems. Based on the lane width of \SI{4}{\m}, an accuracy of \SI{0.75}{\m} is in the following considered as sufficient accuracy as with a typical vehicle width of about \SI{2.20}{\m} the vehicle would stay on its lane. Moreover, only current detections are taken into account. 

\subsection{Convoy Fusion}
\label{subsec:inLane-fusion}
For some scenarios, like driving in a convoy, the distance between the ego vehicle $v_e$ and the cooperative vehicle $v_c$ is less than the sensing range of lane detection systems.

As shown in Fig.~\ref{fig:cld-cases} (a), a cooperative vehicle driving ahead is within the range of the locally detected lane by $v_e$ which leads to an overlap of detected lanes. To determine $l_{\mathrm{coll}}$, the perceived lane information must be split into three parts. The first part is the local perception of the ego vehicle up to the point where the overlap starts. This part is directly transferred to the fused lane. The second part is the overlapping section $l_{\mathrm{ov}}$ of $l_{\mathrm{ego}}$ and $l_{\mathrm{coop}}$. For this section a fusion of the information is necessary; for this purpose different approaches are possible. The accuracy of the lane detection tends to be higher the closer the perception is to the perceiving sensor; thus, we are using a weighted mean fusion.
To achieve a more accurate fusion result, we use the weight $\omega_{\mathrm{coop}}=0.75$ for the information of $v_c$ as this information is closer to the perceiving sensor. However, to not neglect the more precise localization and to compensate possible inaccuracies the information of the ego is incorporated with $\omega_{\mathrm{ego}}=0.25$ into the fusion. Since our assumption is that the detection error is higher than the localization error, we set $\omega_{\mathrm{coop}} > \omega_{\mathrm{ego}}$.
Hence, a point $P$ of $l_{\mathrm{coll}}$ can be determined by Eq.~\eqref{eq:weighted-mean} using two corresponding points from the ego and the cooperative vehicle.
\begin{equation}
    P_{l_{\mathrm{coll}}}=0.25\cdot P_{l_{\mathrm{ego}}}+ 0.75\cdot P_{l_{\mathrm{coop}}}
    \label{eq:weighted-mean}
\end{equation}
In order to complete the construction of $l_{\mathrm{coll}}$, the remaining information from $l_{\mathrm{coop}}$ is taken into account. Besides this approach, we considered a regular mean fusion to construct a globally valid section information for $l_{\mathrm{ov}}$. A regular mean fusion would neglect the fact that the detection of the ego in the overlapping area tends to be less accurate due to the high distance to the sensor and possible partial occlusions by $v_c$. Another approach is to select the information that is expected to be most accurate. In our case, in the range of $l_{\mathrm{ov}}$, the information from the cooperative vehicle would be used, since it is closest to the perceiving sensor. However, this approach would lead to a higher fusion error for an inaccurate lane detection of $v_c$.

It may occur that $l_{\mathrm{ego}}$ and $l_{\mathrm{coop}}$ are offset to each other which would lead to a jump in $l_{\mathrm{coll}}$ at the begin of $l_{\mathrm{ov}}$. For small offsets this could be neglected and the fusion can be performed as mentioned before. Higher offsets require a smooth transition from $l_{\mathrm{ego}}$ to $l_{\mathrm{coop}}$ at $l_{\mathrm{ov}}$. For this purpose a linear interpolation from $l_{\mathrm{ego}}$ to $l_{\mathrm{coop}}$ in the area of $l_{\mathrm{ov}}$ could be applied.

\subsection{Spline Fusion}
\label{subsec:spline-fusion}
In regular traffic without convoy driving, vehicles must ensure a safety distance to the vehicle driving in front which leads to gaps between the lane detection of two vehicles as shown in Fig.~\ref{fig:cld-cases} (b). In this case three to four steps, depending on the road geometry, are necessary to determine the collectively perceived lane $l_{\mathrm{coll}}$. First, it must be determined if a fusion with a cooperative vehicle is plausible and valid. Second, if the vehicles relative positions and their corresponding local detections indicate that a curve is present between $l_{\mathrm{ego}}$ and $l_{\mathrm{coop}}$ an estimation of the apex is necessary. If the lane section appears to be straight, step 2 can be neglected. Third, based on the known points and the estimated apex (if exists) a spline fitting must be performed and finally; fourth, the single parts must be composed to receive $l_{\mathrm{coll}}$. In the following sections we will describe the process of our collective lane detection approach based on these steps.

\subsubsection{Verification of Fusibility}
\label{subsubsec:fusable}
\begin{figure}[h!]
    \centering
    \includegraphics[width=0.95\linewidth, page=2, trim=8.5cm 1.0cm 8.5cm 3.0cm, clip]{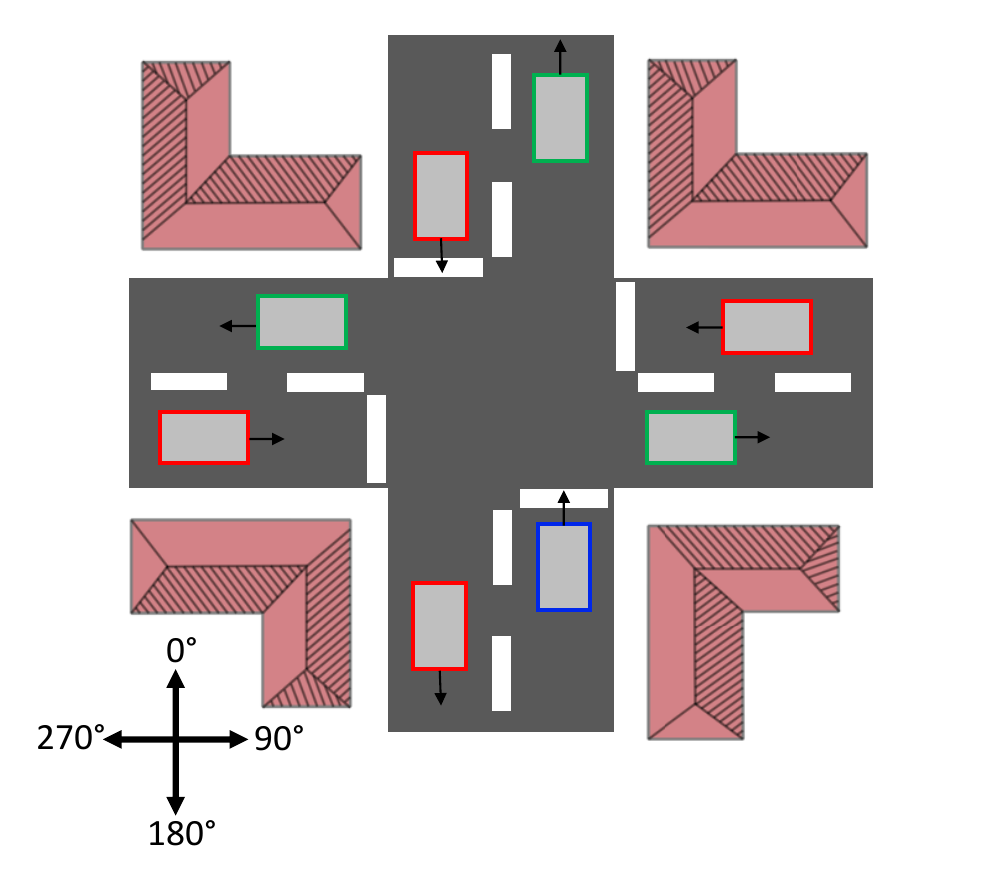}
    \caption{Exemplary intersection scenario to show possible fusion candidates for the ego vehicle (blue). Cooperative vehicles which are on possible trajectories of the ego are considered as valid for fusion (green). Vehicles which are not considered since the resulting lane would not satisfy traffic or trajectory rules are marked red.} 
    \label{fig:fusable}
\end{figure}

In case of a convoy fusion (cf. Sec.~\ref{subsec:inLane-fusion}) it can be easily determined if a fusion is possible. For the spline fusion, it is more difficult. For various scenarios different relative positions for the ego vehicle $v_e$ and a cooperative vehicle $v_c$ occur. As we only want to calculate lanes which are valid and reasonable regarding traffic rules and possible trajectories we must distinguish different cases and filter which cooperative vehicles can be considered as valid for a fusion. An exemplary validation for an intersection scenario is shown in Fig.~\ref{fig:fusable}. The possible motion actions for $v_e$ are driving straight, turning right or turning left. 

We use a left-handed coordinate system to describe the relative position $p_{\mathrm{rel}}(x_{\mathrm{rel}},y_{\mathrm{rel}})$ of $v_c$ to $v_e$. $\Psi_{\mathrm{rel}}$ describes the relative orientation. Considering the scenario shown in Fig.~\ref{fig:fusable} and the possible motions actions, we can impose the following conditions for a cooperative vehicle to be valid for a spline fusion.
\begin{itemize}
    \item ($x_{\mathrm{rel}}>0, y_{\mathrm{rel}}\approx 0, \Psi_{\mathrm{rel}}\approx \SI{0}{\degree})$: $v_c$ driving in front with same direction.
    \item ($x_{\mathrm{rel}}>0, y_{\mathrm{rel}}> 0, \Psi_{\mathrm{rel}}\approx \SI{90}{\degree})$: $v_c$ driving in front and right with direction to the right.
    \item ($x_{\mathrm{rel}}>0, y_{\mathrm{rel}}< 0, \Psi_{\mathrm{rel}}\approx \SI{270}{\degree})$: $v_c$ driving in front and left with direction to the left.
\end{itemize}

\begin{figure*}
     \centering
     \begin{subfigure}[b]{0.48\textwidth}
          \centering
        \includegraphics[width=0.95\linewidth, trim= 0.2cm 0.2cm 0.2cm 0.7cm clip]{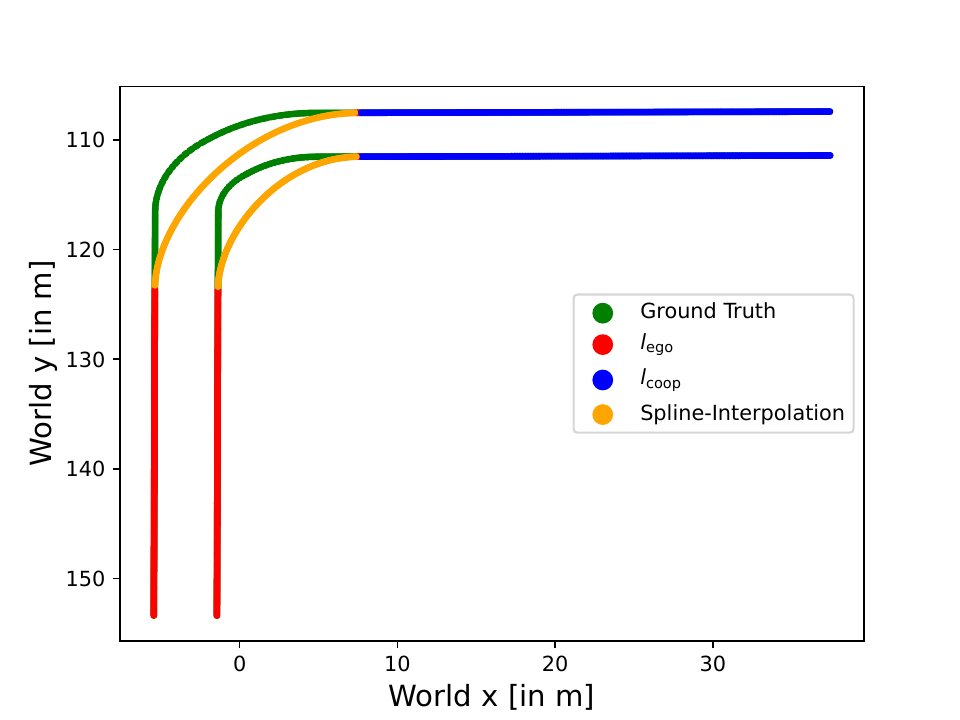}
        \caption{Spline interpolation with error of about \SI{2.4}{\m}.} 
        \label{fig:spline-error}
     \end{subfigure}
     \hfill
     \begin{subfigure}[b]{0.48\textwidth}
         \centering
        \includegraphics[width=0.95\linewidth, trim= 0.2cm 0.2cm 0.2cm 0.7cm clip]{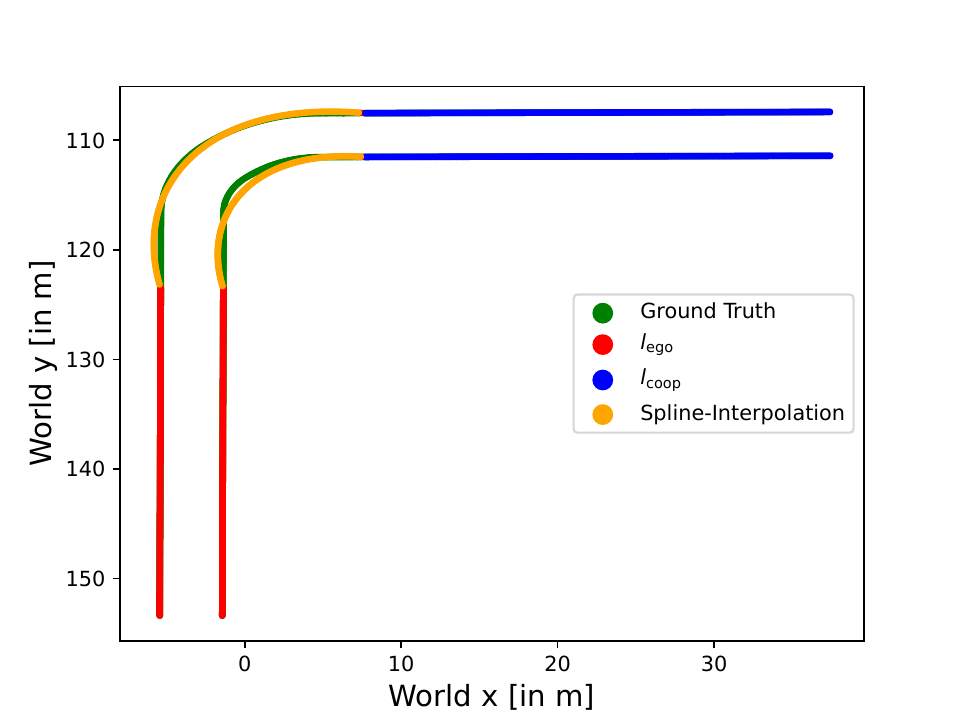}
        \caption{Spline interpolation with error of about \SI{0.4}{\m}.} 
     \label{fig:result-curve}
     \end{subfigure}
     \hfill
        \caption{Spline interpolation of a \SI{90}{\degree} right curve (a) without and (b) with apex estimation}
        \label{fig:apex}
\end{figure*}

For some conditions, single parameters can deviate by a defined threshold (e.g. $y_{\mathrm{rel}}\approx 0$) since the vehicles position can differ inside the lane or even for nearly straight roads a slight curvature can appear which in results in $\Psi_{\mathrm{rel}}\neq 0$. Based on the lane width we defined a positioning deviation threshold $t_{\mathrm{pos}}= \pm \SI{0.40}{\m}$ and an orientation deviation threshold $t_\Psi= \pm \SI{10}{\degree}$.
Similar conditions must be defined for further scenarios, e.g. a highway with multiple lanes in which $y_{\mathrm{rel}}$ should be less than the lane width to avoid a spline fitting over different lanes. Moreover, a maximum interpolation threshold $t_i$ for the spline must be defined. 
If a cooperative vehicle satisfy one of the corresponding conditions and $d_s < d_i$, with $d_s$ as distance between the end of $l_{\mathrm{ego}}$ and the begin of $l_{\mathrm{coop}}$, a fusion can be considered as plausible. 

\subsubsection{Apex Estimation}
\label{subsubsec:apex}
Experiments conducted have shown that with an increasing curve angle, the spline interpolates the curve too flat and which leads so significant errors. This behavior is shown in Fig.~\ref{fig:apex} (a) for a \SI{90}{\degree} right curve. In this case, the interpolation shows an error of about \SI{2.4}{\metre} at the apex of the curve which does not satisfy the requirements since following this estimated lane would lead to leaving the real lane.

To achieve a precise estimation of the unknown lane section, we must enforce that the spline goes through the apex of the curve to prevent the spline from being too flat. Hence, we use a geometrical construction to estimate the apex of the lane boundaries of the curve and consider this estimated point as an additional known point for the spline fitting (cf. Sec.~\ref{subsubsec:spline-fitting}). To perform the apex estimation, all coordinates must be transformed into a common world coordinate system.

\begin{figure}[h!]
    \centering
    \includegraphics[width=0.95\linewidth, page=1, trim= 8cm 4cm 8cm 4cm, clip]{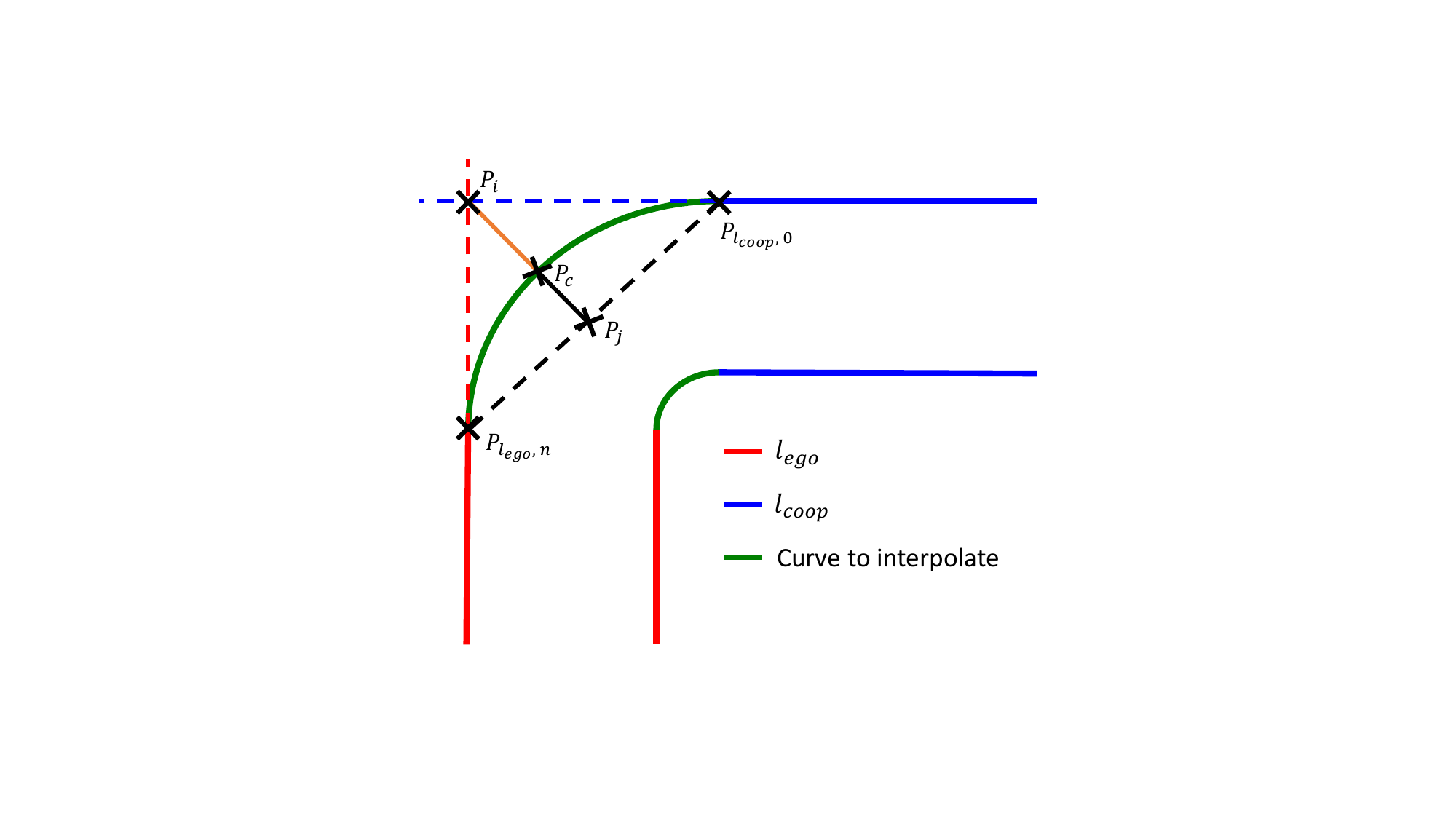}
    \caption{Schematically overview of the geometric estimation of the apex of the left lane boundary. The estimation of the apex for the right lane boundary is performed equally.} 
    \label{fig:apex-estimation}
\end{figure}

We use the geometrical construction presented in Fig.~\ref{fig:apex-estimation}. We construct a triangle containing the arc of the curve between $l_{\mathrm{ego}}$ and $l_{\mathrm{coop}}$.
Therefore, we extend the lines referring to the local lane detection of $v_e$ in driving direction (red dashed line) and the local detection of $v_c$ against driving direction (blue dashed line) until they intersect at point $P_i$. Furthermore, we connect the last point of $l_{\mathrm{ego}}$ ($P_{l_{\mathrm{ego}},n}$) with the first point of $l_{\mathrm{coop}}$ ($P_{l_{\mathrm{coop}}, 0}$) to construct a triangle containing the arc of the curve we need interpolate. Finally, we calculate the orthogonal projection from $P_{l_{\mathrm{ego}},n}P_{l_{\mathrm{coop}}, 0}$ to $P_i$. The point of the projection intersection $P_{l_{\mathrm{ego}},n}P_{l_{\mathrm{coop}}, 0}$ is referred to as $P_j$. Considering the vector $\overrightarrow{P_iP_j}$, the point on the curve can be estimated at about \SI{40}{\percent} of $\overrightarrow{P_iP_j}$. The exact value slightly varies with the curve angle, but experiments have shown that \SI{40}{\percent} leads to the best approximation for different scenarios. Hence, the apex or more generally an additional point on the curve $P_c$ can be determined with Eq.~\eqref{eq:apex}.
\begin{equation}\label{eq:apex}
    P_c=P_i+0.4\cdot \overrightarrow{P_iP_j}
\end{equation}
It must be stated that this estimation is not an exact calculation but as shown in Fig.~\ref{fig:apex} significantly increases the estimation accuracy. 

\subsubsection{Spline Fitting}
\label{subsubsec:spline-fitting}
Generally, a spline is a piecewise polynomial function. For $n+1$ known points ($n \in \mathbb{N}$) $x_0<x_1< \dots < x_n \in \mathbb{R}$ and the corresponding function values $y_0,y_1 \dots,  y_n \in \mathbb{R}$ a spline $S$ is defined as:
\begin{equation}
    \label{eq:spline}
    S:[x_0,x_n]\rightarrow \mathbb{R} \text{ with } S(x_i)=x_i \text{\quad} \forall i\in[0,n],
\end{equation}
where $i\in[0,n-1]$, $s_i=S_{[x_i,x_{i+1}]}$ is a polynom in the interval $[x_i,x_{i+1}]$.

As the spline aims to be a smooth interpolation function for all $i\in[1,n-1]$ the following two conditions must hold:
\begin{enumerate}
    \item $ s'_i(x_i):=s'_{i-1}(x_i)$,
    \item $ s''_i(x_i):=s''_{i-1}(x_i)$.
\end{enumerate}

This leads to the following linear equation system. The parameters $\mu_0,\mu_n,\lambda_0,\lambda_0,b_0$ and $b_n$ depend on the chosen boundary condition.
\begin{equation}
    \label{eq:spline_lgs}
    \begin{aligned}
        \begin{bmatrix}
            \mu_0         & \lambda_0         &                   &                       &                 &        \\
            \frac{h_0}{6} & \frac{h_0+h_1}{3} & \frac{h_1}{6}     &                       &                 &        \\
                          & \ddots            & \ddots            & \ddots                &                 &        \\
                          &                   & \frac{h_{i-1}}{6} & \frac{h_{i-1}+h_i}{3} & \frac{h_{i}}{6} &        \\
                          &                   &                   & \ddots                & \ddots          & \ddots \\
                          &                   &                   &                       & \lambda_n       & \mu_n  \\
        \end{bmatrix} \cdot
        \begin{bmatrix}
            M_0    \\
            M_1    \\
            \vdots \\
            M_i    \\
            \vdots \\
            M_n    \\
        \end{bmatrix}
        \\\
        =
        \begin{bmatrix}
            b_0                                                 \\
            \frac{y_2-y_1}{h_1}-\frac{y_1-y_0}{h_0}             \\
            \vdots                                              \\
            \frac{y_{i+1}-y_i}{h_i}-\frac{y_i-y_{i-1}}{h_{i-1}} \\
            \vdots                                              \\
            b_n                                                 \\
        \end{bmatrix}
    \end{aligned}
\end{equation}

For further information about the linear equation system and the boundary conditions to calculate splines we refer to McKinley and Levine~\cite{mckinley1998cubic}. 

To fit the spline we have to define the known points; as we want to interpolate between the locally detected lane sections $l_{\mathrm{ego}}$ and $l_{\mathrm{coop}}$, these are the available known points. Since the subpolynom between the last data point of $l_{\mathrm{ego}}$ and the first point of  $l_{\mathrm{coop}}$ only depends on the subpolynom before and after, only a few known points are required for the fitting and not all points of $l_{\mathrm{ego}}$ and $l_{\mathrm{coop}}$. We investigated different selections for the known points with \{2, 3, 5, 10\} known points in $l_{\mathrm{ego}}$ and $l_{\mathrm{coop}}$ with \{0, 1, 2, 5\} data points between two known points. These experiments have shown that a higher number of known points could not improve the accuracy of the spline, but the higher number of known points leads to an increased computational effort. Hence, we take two points from each lane boundary of $l_{\mathrm{ego}}$ and $l_{\mathrm{coop}}$. To fit the spline only between the local detections, we take the last and third to last points of the lane boundaries of $l_{\mathrm{ego}}$ as well as the first and the third points from the lane boundaries of $l_{\mathrm{coop}}$ as known points. Furthermore, as described in Sec.~\ref{subsubsec:apex}, an additional known point for the spline fitting is required if the spline should approximate a curve. If the relative orientation of the end of $l_{\mathrm{ego}}$ and the begin of $l_{\mathrm{coop}}$ is $> \SI{10}{\degree}$ we assume a curve and estimate the apex as an additional fifth point per lane boundary to fit the spline.
Each locally detected lane is in the corresponding coordinate system of the vehicle which perceived this section. To fit the spline the determined known points must be transformed to a common coordinate system. Hence, we transform all coordinates to the world frame with the assumption that this coordinate transformation is error free. 

After the transformation of all known points the linear equation system (cf. Eq.~\eqref{eq:spline_lgs}) describing the fitting problem can be solved to obtain the spline interpolating the gap between $l_{\mathrm{ego}}$ and $l_{\mathrm{coop}}$.
 
\subsubsection{Final Lane Construction}
\label{subsubsec:construction}
After the spline fitting we construct $l_{\mathrm{coll}}$ based on $l_{\mathrm{ego}}$ and $l_{\mathrm{coop}}$ as well as the spline $S$ to interpolate between the local detection. Therefore, we use $l_{\mathrm{ego}}$ as begin of $l_{\mathrm{coll}}$. At the end of $l_{\mathrm{ego}}$, we sample points of $S$ with a distance of \SI{0.10}{\metre} to transfer the information from the spline to the Lanelet format. Afterwards, we concatenate these points at $l_{\mathrm{ego}}$. At the begin of $l_{\mathrm{coop}}$, we have a local detection again and no more interpolation is necessary. As for the convoy fusion, we use the locally detected lane of $v_c$ and append this information to the concatenation of $l_{\mathrm{ego}}$ and the sampled points of $S$. 

\subsection{Information Exchange}
\label{subsec:exchange}
For CoLD we should not only consider the perception side but also the communication. As we set our focus to the enhancements in perception, the communication will be only discussed briefly. For our experiments we assume ideal conditions for the V2X communication without any bandwidth limitations or communication delays. Hence, we use highly accurate sampled lanelets as data format to exchange the locally perceived lanes at each perception iteration. Considering a perceived lane with a length of \SI{30}{\metre}, this leads to two lane boundaries with 300 data points each, leading to 7200 bytes per perceived lane. For multiple vehicles perceiving not only a single lane but multiple lanes this would lead to a overload of the V2X communication channel. Thus, the Lanelet format is not a suitable format to exchange the lane information. As a format for information exchange, the Beziér-Spline appears to be well suited. As shown in Sec.~\ref{sec:related_work}, splines are capable to model lane sections. Moreover, a Beziér-Spline is defined by start point, end point, and two additional control points to form the spline. Using this format an efficient and accurate information exchange with 96 bytes per lane section, independent from the length, is possible.

Furthermore, the message frequency should be considered as there is no necessity to exchange the information each perception iteration as the lane is static. The ETSI defined a message format and generation rules for the information exchange of dynamic objects which can not be applied to lanes. We suggest an exchange rate depending on the perception range and the current velocity of the vehicle. Determining exact frequency suggestions will be part of future work.


\section{RESULTS}
\label{sec:results}

For the evaluation of the proposed approach we use the CARLA simulator by Dosovitskiy et al.~\cite{CARLA}. For sensor data processing, data fusion and evaluation the RESIST framework of M{\"u}ller et al.~\cite{resist} including the extensions from Volk et al.~\cite{volk_environment-aware_2019,volk2021} is employed. 
First, we present the real-time capability of our system by a runtime analysis. Afterwards, we show the performance of our collective lane detection based on three scenarios (straight lane, \SI{90}{\degree} right curve, and a rural road) with two vehicles. The evaluation is performed on a system with an AMD Ryzen 7 5700X @ \SI{3.40}{\giga\hertz}, 32 GB RAM and a Nvidia RTX3080 GPU. 

For development purposes we use a generic lane sensor for vehicle-local lane detection. Using state of the art lane detectors could lead to uncontrollable errors and furthermore, most publications only evaluate their algorithms based on performance metrics like precision and recall but do not provide any information about the perception error in \si{\metre}. Moreover, a generic lane sensor allows a detailed evaluation using different error models such as offset or noise. The generic lane sensor is implemented in the C\texttt{++} CARLA API of the RESIST framework and uses the underlying OpenDrive Map to generate data points of each lane boundary with a distance of \SI{0.10}{\metre} between two points. Each lane is then represented by two lists of points and corresponds to the Lanelet format~\cite{lanelet} which we use to maintain our lane information. For our experiments we set the perception range of our lane detection sensor to $r_{\mathrm{lane}}=\SI{30}{\metre}$. Moreover, the functionality of the Lanelet2 framework~\cite{lanelet2} is used to read the map transformed by the OpenDrive2Lanelet converter~\cite{opendrive2lanelet} as ground truth (GT) and the matching module to extract the corresponding GT for the detected lanes out of the GT map. For the spline fitting we use the calculation provided by Aly~\cite{aly2008real}.

\subsection{Runtime Analysis}
\label{subsec:runtime}
To show the real-time capability of our approach we examine both fusion types on the rural scenario with 300 frames. The results are presented in Tab.~\ref{tab:runtime}.
\begin{table}[H]
    \centering
    \begin{tabular}{lccc }
    \toprule
     & mean [\si{\milli\second}] & $\sigma$ [\si{\milli\second}] & max [\si{\milli\second}] \\
    \midrule
    Convoy Fusion & 2.23 & 0.42 & 3.66\\
    Spline Fusion& 3.62 & 0.70 & 6.62 \\
    \bottomrule
    \end{tabular}
    \caption{Runtime analysis for CoLD fusion. Results in~\si{\milli\second}.}
    \label{tab:runtime}
\end{table}

\begin{table*}[t]
    \centering
    \begin{tabular}{lr ccc }
    \toprule
    & & & \textbf{Convoy Fusion} & \textbf{Spline Fusion} \\

    \textbf{Road type}& & $l_{\mathrm{ego}}$ &  $l_{\mathrm{coll}}$ &  $l_{\mathrm{coll}}$   \\
    \midrule
    \multirow{ 2}{*}{\textbf{Straight Lane}} & MSE (left / right)~[\si{\metre}]              &  (0.005 / 0.005)  & (0.005 / 0.005) & (0.005 / 0.005)    \\ 
    & MAX (left / right)~[\si{\metre}]             &  (0.010 / 0.011)  & (0.010 / 0.011) & (0.027 / 0.028)    \\
    \midrule
    \multirow{ 2}{*}{\textbf{Right Curve} (\SI{90}{\degree})}& MSE (left / right)~[\si{\metre}]              &  (0.005 / 0.005)  & (0.005 / 0.005) & (0.029 / 0.030)    \\ 
    & MAX (left / right)~[\si{\metre}]             &  (0.010 / 0.011)  & (0.010 / 0.011) & (0.693 / 0.707)    \\
    \midrule
    \midrule
    \multirow{ 2}{*}{\textbf{Rural Road}}& MSE (left / right)~[\si{\metre}]              &  (0.005 / 0.005)  & (0.005 / 0.005) & (0.023 / 0.020)    \\ 
    & MAX (left / right)~[\si{\metre}]             &  (0.010 / 0.010)  & (0.010 / 0.010) & (0.560 / 0.572)    \\
    \midrule
    \midrule
    \addlinespace
    & Perception range~[\si{\metre}]      &  30.00  &  55.00 & 80.00 - 90.00\\
    \addlinespace
    \bottomrule
    \end{tabular}
    \caption{Results for local perception of an ego vehicle ($l_{\mathrm{ego}}$) and the collective perception ($l_{\mathrm{coll}}$) for convoy and spline fusion. The perception range depends on the maximum valid interpolation distance for the given scenario. Results in~\si{\metre}.}
    \label{tab:results}
\end{table*}
For both fusion types it must be evaluated if an overlap of the lanes exist which takes about \SI{0.33}{\ms}.
For the convoy fusion we observed a mean runtime of \SI{2.23}{\ms} including matching, fusion and concatenation of the information. For the spline fusion the fusibility check (cf. Sec.~\ref{subsubsec:fusable}) is performed in about \SI{0.01}{\ms}. Including spline fitting, sampling and the construction of $l_{\mathrm{coll}}$ we observed a mean runtime of \SI{3.62}{\ms}.

Considering the sensor frequency of $\SI{100}{\ms}$ as real-time threshold, both fusion types achieve runtimes far below even for their maximum runtime. 
\subsection{Perception Evaluation}
\label{subsec:perception-results}

For the evaluation of our CoLD approach we consider the distance between the collectively perceived lane and the GT. For performance evaluation, we consider the mean squared error $\mathrm{mse}_l$ for $l_{\mathrm{coll}}$ with $n$ points as shown in Eq.~\eqref{eq:mse}.

\begin{equation}\label{eq:mse}
    \mathrm{mse}_l=\frac{1}{n}\sum_{i=1}^n |P_s - P_{GT}|_i
\end{equation}

$P_s$ refers to a point of $l_{\mathrm{coll}}$ and $P_{GT}$ to the corresponding GT point. As $P_{GT}$ is determined by a closest point matching the resulting error depends on the sampling rate of the GT. We set a sampling distance of \SI{0.02}{\m}, as this corresponds to the maximum accuracy of the CARLA waypoint information.
Regarding safety aspects higher deviations of the perceived lane are more relevant. Hence, we are evaluating not only $\mathrm{mse}_l$ but also the maximum deviation. As single outliers can occur and be filtered out to not affect motion planning negatively, we consider the 95th percentile of the error in lane detection to evaluate the safety.

An overview of the observed perception errors and ranges is presented in Tab.~\ref{tab:results}.

The vehicle-local perception of the ego achieved an error of \SIrange{5}{10}{\mm} which can be attributed to the generic lane sensor used. The sensor uses the GT as detected lane; thus, only a minimal deviation through the evaluation process occurs. The small error appears due to the aforementioned sampling rate of the GT.

For the convoy fusion we can observed that the perception error of the collectively perceived lane is equal to the local detection error for all test scenarios. As we construct $l_{\mathrm{coll}}$ solely out of $l_{\mathrm{ego}}$ and $l_{\mathrm{coop}}$, the local perception errors are transferred to the collective perception. However, this experiment shows that the weighted mean fusion does not affect the perception accuracy in terms of MSE and maximum error negatively. With this observation it can be stated that the error of the collective perception does not exceed the local perception error. Thus, under the assumptions made in Sec.~\ref{sec:method} the CP will satisfy the accuracy requirements if the local perception with real-world systems achieves a sufficient accuracy. Moreover, the convoy fusion is robust against minor noise or offset errors of the local detection as the error does not increase through the fusion. With a distance $d_v=\SI{25}{\m}$ between $v_e$ and $v_c$ we achieve a total perception range of about \SI{55}{\m}. As $d_v$ is not constant due to varying vehicle velocities small variations occur; hence, the total perception range is not exactly \SI{55}{\m} at each perception iteration.

In contrast to the convoy fusion, the spline fusion lead to higher errors. For a scenario with a straight lane like for highways, we observed a very accurate estimation. For $d_s=\SI{30}{\m}$ the mean error was \SIrange{5}{10}{\mm}, the maximum error increases up to \SI{0.03}{\metre} as the spline is not exactly straight based on small deviations of the known points. However, even for higher interpolation distances up to \SI{50}{\metre} the maximum deviation of $l_{\mathrm{coll}}$ was about \SI{0.05}{\metre}. Considering the perception range we achieve an extension from \SI{30}{\metre} up to \SI{90}{\metre} with $d_s=\SI{30}{\metre}$ and \SI{110}{\metre} with $d_s=\SI{50}{\metre}$. As for the convoy fusion it must be stated that the total perception range can slightly variate due to non constant velocities.

For a \SI{90}{\degree} curve with $d_s=\SI{20}{\m}$ we achieved a $\mathrm{mse}_l$ of about \SI{0.03}{\m}. The maximum deviation for the right lane boundary was \SI{0.70}{\m}. The low MSE in contrast to the high maximum error can be attributed to the scenario as it contains straight sections before and after the curve. However, as shown in Fig.~\ref{fig:apex} (b) following the estimated lane would not lead to leaving the lane; further, the shape of the curve is modelled with a sufficient accuracy in comparison to the approximation without apex estimation (cf. Fig.~\ref{fig:apex} (a)). Increasing $d_s$ leads to higher errors which could lead to leaving the lane and hence, do not satisfy safety constraints. If the distance to interpolate is smaller, the accuracy increases significantly.

\begin{figure}[h!]
    \centering
    \includegraphics[width=\linewidth, trim= 3cm 1.2cm 5.2cm 1.5cm clip]{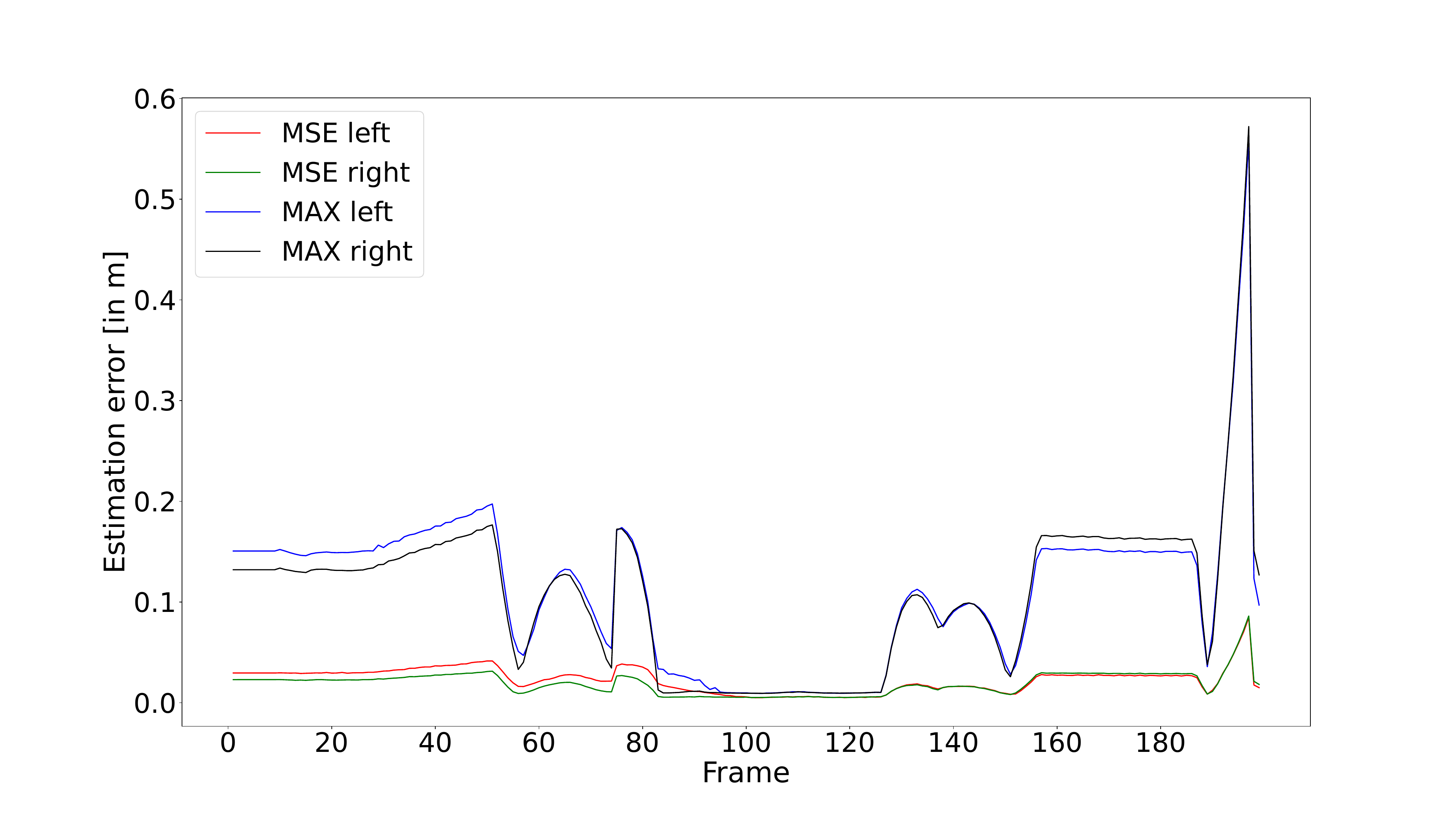}
    \caption{Estimation error of CoLD with spline fusion in rural scenario. Results in \si{\m}.} 
    \label{fig:result-rural}
\end{figure}

\begin{figure}[h!]
    \centering
    \includegraphics[width=0.95\linewidth, trim= 0.2cm 0.2cm 0.2cm 0.7cm clip]{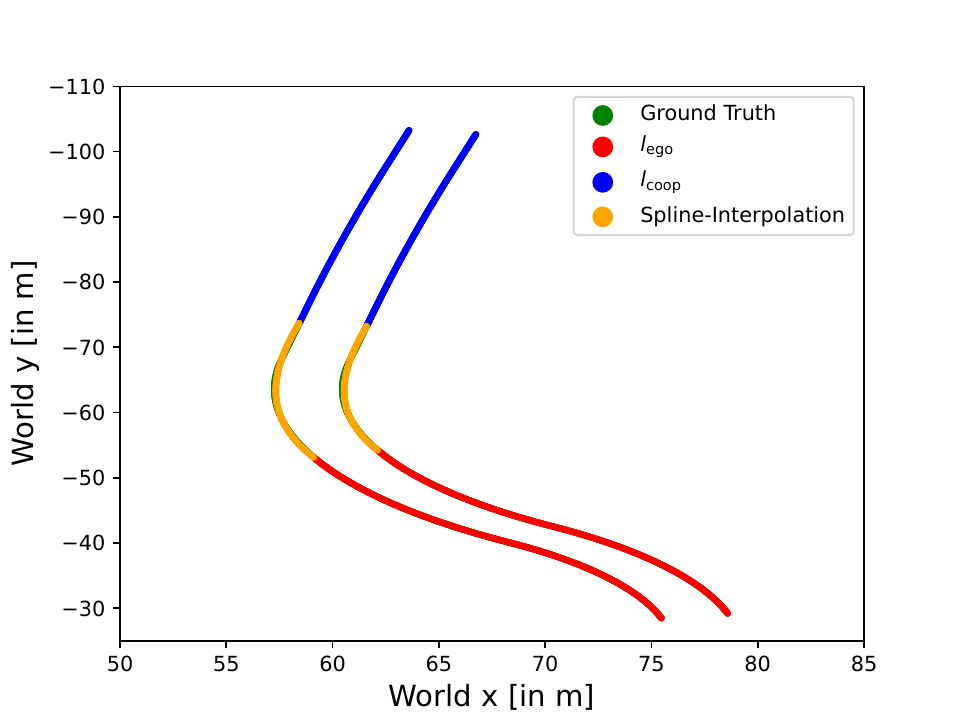}
    \caption{CoLD with spline fusion in rural scenario with a maximum error at the apex of the curve of about \SI{0.15}{\m}.} 
    \label{fig:result-lanelet-rural}
\end{figure}

Considering a rural road scenario with $d_s = \SI{20}{\m}$, we can observe a $\mathrm{mse}_l$ of about \SI{0.02}{\m} which can be considered as highly accurate. The maximum error over all frames of the scenario was about \SI{0.57}{\m}. However, this significant deviation only appeared for a single frame (cf. Fig.~\ref{fig:result-rural}) at the transition from a curve to a straight section. Further frames have an error of mostly less than \SI{0.20}{\m}. The maximum error averaged over all frames is about \SI{0.11}{\m} which can be considered as accurate. 
Figure~\ref{fig:result-lanelet-rural} shows an exemplary estimation for the rural scenario at frame 65; even for an estimation error of about \SI{0.15}{\m}, the shape of the approximated lane is similar to the real lane.
With $d_s=\SI{20}{\m}$ and $r_{\mathrm{lane}}=\SI{30}{\metre}$ a total perception range of about \SI{80}{\m} can be achieved. For higher interpolation distances ($\geq$ \SI{30}{\m}) the maximum error exceeds \SI{1}{\m}; hence, to achieve a reliable accuracy the maximum interpolation distance for this scenario is about \SI{20}{\m}.

As for the convoy fusion, the fusion with spline interpolation appears to be robust against offset errors in the local detection. An offset in the local detection leads to an offset in the spline but not significantly higher than the local offset. However, the spline interpolation is not robust against noise on the local detection. If the known points for the spline fitting deviate in different directions or indicate a wrong direction of the shape of the lane section, errors of several meters can appear or moreover, a spline fitting is not possible anymore. However, it must be stated that the collective perception for lanes as well as for traffic participants requires a sufficient accuracy of the local perception to achieve a contribution.

The results of the spline interpolation heavily depend on the distance to interpolate and the road geometry. However, for long-distance interpolation, the road layout cannot be reconstructed safely anymore. Thus, a maximum interpolation distance of about \SI{20}{\m} is suggested, as this corresponds to a distance with sufficient accuracy in all evaluated scenarios.

Moreover, it must be stated that the maximum deviations for the \SI{90}{\degree} curve and the rural scenario do not satisfy the accuracy requirements for autonomous driving. However, the shape of the lane is modelled with a good accuracy to make a contribution to the motion planning as the extended perception range allows to estimate if e.g. a curve follows the local detection and thus, the speed of the vehicle should be adapted. Furthermore, the motion planning can be improved with further vehicle-local perception iterations, based on the extended sensing this requires only minor adaptions.

\section{CONCLUSION \& OUTLOOK}
\label{sec:conclusion}

Within the scope of this paper we presented a real-time capable approach for collective lane detection. The system aims to extend the sensing range for lane detection by using information from distributed cooperative vehicles to enable a more detailed motion planning.
As long-distance fusion, the environment (curves, obstacles) between the vehicles cannot be reconstructed safely anymore we focused on short (convoy driving) and medium-distance collective perception. In convoy driving scenarios we use a concatenation of local detections to extend the sensing range by up to \SI{100}{\percent}, achieving an accuracy of \SI{0.01}{\m} depending on the local lane detection. For more distant vehicles a gap between local detections appears which requires an estimation of this lane section. For this purpose, we used a spline to approximate the lane by interpolation between the locally perceived lane sections. To improve the interpolation we proposed an apex estimation of the lane in curvy scenarios. The perception range could be extended by up to \SI{200}{\percent} in contrast to the local detection. For all three simulated scenarios we achieved a sufficient accuracy for motion planning with a maximum deviation of \SI{0.03}{\m} for a straight lane, \SI{0.70}{\m} for a \SI{90}{\degree} curve and \SI{0.57}{\m} for a rural road.
Experiments conducted to evaluate the runtime showed a mean result of \SIrange{3}{4}{\ms} which can be considered as real-time capability.

As a next step, various state of the art lane detection algorithms will be investigated to verify the robustness of the fusion approach under various conditions with more vehicles. Additionally, the estimation of the further known point for the spline interpolation will be optimized based on an approximation of the curve angle.
For further research also possible message generation rules will be investigated. Moreover, it will be considered if transmitting previously perceived information could improve the accuracy of a collective lane detection.






\section*{ACKNOWLEDGMENT}

This work has been partially funded by the German Research Foundation
(DFG) in the priority program 1835 under grant BR2321/5-2.


\bibliographystyle{IEEEtran} 
\bibliography{literature.bib}

@inproceedings{CARLA,
  title="{CARLA: An Open Urban Driving Simulator}",
  author={Dosovitskiy, Alexey and Ros, German and Codevilla, Felipe and Lopez, Antonio and Koltun, Vladlen},
  booktitle={Conference on robot learning},
  pages={1--16},
  year={2017},
  organization={PMLR}
}

@inproceedings{resist,
  title={{Framework for Varied Sensor Perception in Virtual Prototypes}},
  author={Mueller, Stefan and Hospach, Dennis and Gerlach, Joachim and Bringmann, Oliver and Rosenstiel, Wolfgang},
  booktitle={MBMV},
  pages={145--154},
  year={2015}
}

@INPROCEEDINGS{volk2021,
  author={Volk, Georg and Delooz, Quentin and Schiegg, Florian A. and Von Bernuth, Alexander and Festag, Andreas and Bringmann, Oliver},
  booktitle={2021 IEEE International Intelligent Transportation Systems Conference (ITSC)}, 
  title="{Towards Realistic Evaluation of Collective Perception for Connected and Automated Driving}", 
  year={2021},
  volume={},
  number={},
  pages={1049-1056},
  doi={10.1109/ITSC48978.2021.9564783}}

@inproceedings{volk_environment-aware_2019,
	address = {Paris, France},
	title = {Environment-aware {Development} of {Robust} {Vision}-based {Cooperative} {Perception} {Systems}},
	isbn = {978-1-72810-560-4},
	doi = {10.1109/IVS.2019.8814148},
	language = {en},
	urldate = {2019-10-02},
	booktitle = {2019 {IEEE} {Intelligent} {Vehicles} {Symposium} ({IV})},
	publisher = {IEEE},
	author = {Volk, Georg and von Bemuth, Alexander and Bringmann, Oliver},
	month = jun,
	year = {2019},
	pages = {126--133},
}

@inproceedings{volk2019towards,
title={{Towards Robust CNN-based Object Detection through Augmentation with Synthetic Rain Variations}},
ISSN={null},
DOI={10.1109/ITSC.2019.8917269},
booktitle={2019 IEEE Intelligent Transportation Systems Conference (ITSC)},
author={Volk, Georg and M\"uller, Stefan and Bernuth, Alexander von and Hospach, Dennis and Bringmann, Oliver},
year={2019},
month={Oct},
pages={285-292}
}

@misc{europeanUnion2019,
  author       = {{European Commission}},
  title        = "{Road safety: Commission elcomes agreement on new EU rules to help
save lives}",
  year         = {2019},
  note      = {[{O}nline]. {Available}: \url{https://ec.europa.eu/commission/presscorner/detail/en/IP\_19\_1793}, [Accessed: December 5, 2022]}

}

@inproceedings{rauch2012,
  title="{Car2X-Based Perception in a High-Level Fusion Architecture for Cooperative Perception Systems}",
  author={Rauch, Andreas and Klanner, Felix and Rasshofer, Ralph and Dietmayer, Klaus},
  booktitle={2012 IEEE Intelligent Vehicles Symposium},
  pages={270--275},
  year={2012},
  organization={IEEE}
}

@misc{ETSI-CAM,
   title = {{Intelligent Transport Systems (ITS); Vehicular Communications; Basic Set of Applications; Part 2: Specification of Cooperative Awareness Basic Service}},
   author = {{ETSI}},
   year = {2019},
   note = {{ETSI EN 302 637-2 V1.4.1}},
   month = 04
   }

@inproceedings{schiegg2020,
  title="{Analytical Performance Evaluation of the Collective Perception Service in IEEE 802.11p Networks}",
  author={Schiegg, Florian A and Bischoff, Daniel and Krost, Johannes R and Llatser, Ignacio},
  booktitle={2020 IEEE Wireless Communications and Networking Conference (WCNC)},
  pages={1--6},
  year={2020},
  organization={IEEE}
}

@inproceedings{delooz2020,
  title="{Revisiting Message Generation Strategies for Collective Perception in Connected and Automated Driving}",
  author={Delooz, Quentin and Festag, Andreas and Vinel, Alexey},
  booktitle={VEHICULAR 2020: The Ninth International Conference on Advances in Vehicular Systems, Technologies and Applications, Porto, Portugal, 18-22 April, 2020},
  pages={46--52},
  year={2020},
  organization={International Academy, Research and Industry Association (IARIA)}
}

@inproceedings{thandavarayan2019,
  title="{Analysis of Message Generation Rules for Collective Perception in Connected and Automated Driving}",
  author={Thandavarayan, Gokulnath and Sepulcre, Miguel and Gozalvez, Javier},
  booktitle={2019 IEEE Intelligent Vehicles Symposium (IV)},
  pages={134--139},
  year={2019},
  organization={IEEE}
}

@INPROCEEDINGS{lanelet,  
author={Bender, Philipp and Ziegler, Julius and Stiller, Christoph},  
booktitle={2014 IEEE Intelligent Vehicles Symposium Proceedings},   
title="{Lanelets: Efficient Map Representation for Autonomous Driving}",   
year={2014},  
volume={},  
number={},  
pages={420-425},  
doi={10.1109/IVS.2014.6856487}
}

@inproceedings{lanelet2,
  title     = "{Lanelet2: A High-Definition Map Framework for the Future of Automated Driving}",
  author    = {Poggenhans, Fabian and Pauls, Jan-Hendrik and Janosovits, Johannes and Orf, Stefan and Naumann, Maximilian and Kuhnt, Florian and Mayr, Matthias},
  booktitle = {Proc.\ IEEE Intell.\ Trans.\ Syst.\ Conf.},
  year      = {2018},
  address   = {Hawaii, USA},
  month     = {November}
}

@INPROCEEDINGS{opendrive2lanelet,  
author={Althoff, Matthias and Urban, Stefan and Koschi, Markus},  
booktitle={2018 IEEE International Conference on Service Operations and Logistics, and Informatics (SOLI)},   
title="{Automatic Conversion of Road Networks from OpenDRIVE to Lanelets}",   
year={2018},  
volume={}, 
number={},  
pages={157-162},  
doi={10.1109/SOLI.2018.8476801}
}

@inproceedings{CommonRoad,
	author = {Althoff, Matthias and Koschi, Markus and Manzinger, Stefanie},
	title = "{CommonRoad: Composable Benchmarks for Motion Planning on Roads}",
	booktitle = {Proc. of the IEEE Intelligent Vehicles Symposium},
	year = {2017}
}

@inproceedings{dupuis2010opendrive,
  title="{OpenDRIVE 2010 and Beyond--Status and Future of the de facto Standard for the Description of Road Networks}",
  author={Dupuis, Marius and Strobl, Martin and Grezlikowski, Hans},
  booktitle={Proc. of the Driving Simulation Conference Europe},
  pages={231--242},
  year={2010}
}

@article{haklay2008openstreetmap,
  title="{Openstreetmap: User-generated Street Maps}",
  author={Haklay, Mordechai and Weber, Patrick},
  journal={IEEE Pervasive computing},
  volume={7},
  number={4},
  pages={12--18},
  year={2008},
  publisher={Ieee}
}

@article{wang2000lane,
  title="{Lane Detection Using Spline Model}",
  author={Wang, Yue and Shen, Dinggang and Teoh, Eam Khwang},
  journal={Pattern Recognition Letters},
  volume={21},
  number={8},
  pages={677--689},
  year={2000},
  publisher={Elsevier}
}

@inproceedings{zhao2012novel,
  title="{A Novel Multi-Lane Detection and Tracking System}",
  author={Zhao, Kun and Meuter, Mirko and Nunn, Christian and M{\"u}ller, Dennis and M{\"u}ller-Schneiders, Stefan and Pauli, Josef},
  booktitle={2012 IEEE Intelligent Vehicles Symposium},
  pages={1084--1089},
  year={2012},
  organization={IEEE}
}

@inproceedings{aly2008real,
  title="{Real Time Detection of Lane Markers in Urban Streets}",
  author={Aly, Mohamed},
  booktitle={2008 IEEE intelligent vehicles symposium},
  pages={7--12},
  year={2008},
  organization={IEEE}
}

@inproceedings{pan2018SCNN,
 author = {Pan, Xingang and Shi, Jianping and Luo, Ping and Wang, Xiaogang and Tang, Xiaoou},
 title = "{Spatial As Deep: Spatial CNN for Traffic Scene Understanding}",
 booktitle = {AAAI Conference on Artificial Intelligence (AAAI)},
 month = {February},
 year = {2018} 
}

@article{zou2019robust,
  title="{Robust Lane Detection from Continuous Driving Scenes Using Deep Neural Networks}",
  author={Zou, Qin and Jiang, Hanwen and Dai, Qiyu and Yue, Yuanhao and Chen, Long and Wang, Qian},
  journal={IEEE transactions on vehicular technology},
  volume={69},
  number={1},
  pages={41--54},
  year={2019},
  publisher={IEEE}
}

@article{mamun2022comprehensive,
  title="{A Comprehensive Review on Lane Marking Detection Using Deep Neural Networks}",
  author={Mamun, Abdullah Al and Ping, Em Poh and Hossen, Jakir and Tahabilder, Anik and Jahan, Busrat},
  journal={Sensors},
  volume={22},
  number={19},
  pages={7682},
  year={2022},
  publisher={MDPI}
}

@article{mckinley1998cubic,
  title={Cubic Spline Interpolation},
  author={McKinley, Sky and Levine, Megan},
  journal={College of the Redwoods},
  volume={45},
  number={1},
  pages={1049--1060},
  year={1998}
}

@article{runge1901empirische,
  title={{\"U}ber empirische Funktionen und die Interpolation zwischen {\"a}quidistanten Ordinaten},
  author={Runge, Carl},
  journal={Zeitschrift f{\"u}r Mathematik und Physik},
  volume={46},
  number={224-243},
  pages={20},
  year={1901}
}

\end{document}